\documentclass[letterpaper]{article} 
\usepackage{aaai24}  
\usepackage{times}  
\usepackage{helvet}  
\usepackage{courier}  
\usepackage[hyphens]{url}  
\usepackage{graphicx} 
\usepackage{tabularx} 
\usepackage{array}
\usepackage{natbib}  
\usepackage{caption} 
\usepackage{algorithm}
\usepackage{algorithmic}
\usepackage{xcolor}
\usepackage{newfloat}
\usepackage{listings}
\DeclareCaptionStyle{ruled}{labelfont=normalfont,labelsep=colon,strut=off} 
\floatstyle{ruled}
\newfloat{listing}{tb}{lst}{}
\floatname{listing}{Listing}
\title{Procedural Content Generation in Games: A Survey with Insights on Emerging LLM Integration}
\author{
    Mahdi Farrokhi Maleki,
    Richard Zhao
}
\affiliations{
    Department of Computer Science, University of Calgary\\
    Calgary, Alberta, Canada, T2N 1N4\\
    mahdi.farrokhimaleki@ucalgary.ca, richard.zhao1@ucalgary.ca
}

\begin{document}

\maketitle

\begin{abstract}
Procedural Content Generation (PCG) is defined as the automatic creation of game content using algorithms. PCG has a long history in both the game industry and the academic world. It can increase player engagement and ease the work of game designers. While recent advances in deep learning approaches in PCG have enabled researchers and practitioners to create more sophisticated content, it is the arrival of Large Language Models (LLMs) that truly disrupted the trajectory of PCG advancement.

This survey explores the differences between various algorithms used for PCG, including search-based methods, machine learning-based methods, other frequently used methods (e.g., noise functions), and the newcomer, LLMs. We also provide a detailed discussion on combined methods. Furthermore, we compare these methods based on the type of content they generate and the publication dates of their respective papers. Finally, we identify gaps in the existing academic work and suggest possible directions for future research.

\end{abstract}

\section{Introduction}
The video game industry has been expanding rapidly and even surpassed the combined revenue of the music and movie industries in 2022 \cite{Arora_2023}. This huge market means there is always a need for new content. However, the process of creating a game is very time-consuming and can take several years \cite{Juego_Studio_2024}. Procedural Content Generation (PCG) is considered one of the solutions to this problem and can increase replay value, reduce production costs, and minimize effort \cite{amato2017procedural}. PCG for games has existed since the 1980s, and it was mainly used in roguelike games such as \textit{Beneath Apple Manor} (1978) and the genre's namesake, \textit{Rogue} \cite{rogue1980}.

PCG can be used to create a variety of content, but it is commonly used to create art assets \cite{kang2020image, mittermueller2022est}, maps and levels \cite{kreitzer2019automatic,kumaran2019generating}, game mechanics \cite{machado2019evaluation}, and music for games \cite{makhmutov2019adaptive}. The algorithms used can greatly vary depending on the content they are supposed to generate, but we can generally categorize them under a few categories. Most of the algorithms reviewed in this survey fall under one or a combination of three categories: (1) search-based methods, such as Monte Carlo Tree Search (MCTS), which focus mainly on optimizations, (2) learning-based methods, including traditional machine learning and deep learning (DL), such as generative adversarial networks (GAN) and reinforcement learning (RL), which are the recent additions to PCG, and (3) other methods, such as noise functions and generative grammars, that we could not categorize in the earlier categories, Because of the massive interest in DL in recent years, there are many papers published that use DL algorithms as part of PCG. The newest player on the team, Large Language Models (LLMs), comes with very unique characteristics that we will describe in detail.

\subsection{Related Works}
There are a handful of surveys on PCG for games with different focuses and aims that have been published before our work. Some of them discuss the technical details of PCG algorithms \cite{zhang2022survey}, while others focus on content created by PCG \cite{hendrikx2013procedural,de2011survey}. Some papers focus on specific types of PCG; for example, machine learning in PCG, while other works \cite{summerville2018procedural} mainly focus on DL algorithms in PCG.  Liu et al. (\citeyear{liu2021deep}) specifically examines puzzle generation using PCG. Two textbooks, one for PCG \cite{shaker2016procedural} and one for Game AI \cite{yannakakis2018artificial}, cover search-based methods, solver-based methods, constructive generation methods (such as cellular automata and grammar-based methods), fractals, noise, and ad-hoc methods for generating diverse game content.

As of the writing of this paper, the most recent survey on PCG was published in 2022 \cite{zhang2022survey}. This paper includes different categories of PCG; however, it does not discuss generated content in detail nor mention gaps in the field or possible future directions. Based on this information, it is clear that there is a lack of a comprehensive review of PCG for games in recent years. Since the publication of Liu et al. (\citeyear{liu2021deep}), PCG has grown quickly, and a significant number of papers and articles, especially those discussing DL and LLMs, have been published. New research directions and possibilities have emerged in ways that significantly changed the research field. A review of the state-of-the-art and the latest applications of DL and LLMs to PCG is needed.

\subsection{Review Structure}
The structure of the paper is as follows:

First, we give an overview of the different types of content that can be generated using PCG. Then, we go through each category and explain the most commonly used methods published in academic articles. While some cutting-edge deep learning methods are applied on their own, others are applied in combination with more traditional methods, or in an interactive setting \cite{liu2021deep}. Because of this, we added another category called combined methods. The reason behind this choice is that we saw a new trend in combining different algorithms (especially GANs, RL, and LLMs), and we believe that this new category could shape the future of PCG for games. Following this, using the data we gathered by analyzing recently published papers, we discuss the research trends, areas of focus in PCG, gaps in academic research, and possible solutions. We end the paper with an outlook on possible future directions and a conclusion. 

For reviewing the mentioned articles, we had several criteria for inclusion and exclusion, which we explain below:
\begin{enumerate}
    \item To ensure a deeper focus on recently published papers, we selected related papers on PCG for games published in one of five different well-known conference series: the Artificial Intelligence and  Interactive Digital Entertainment (AIIDE) conference series, the International Conference on the Foundations of Digital Games (FDG) series, the Advances in Computer Games (ACG) conference series, the Computer-Human Interaction in Play (CHI-PLAY) conference series, and the IEEE Conference on Games (CoG) series, during the most recent five years (2019-2023). Based on these criteria, we obtained 207 articles aggregated in Figure 1. 
    The distribution of selected papers is as follows: FDG: 40\%, CoG: 33\%, AIIDE: 24\%, CHI PLAY: 2\%, ACG: 1\%.
    
For works before this period, we only discuss influential or interesting papers with a high citation count in the mentioned survey. For the latter, we used Google Scholar for academic papers and Google for grey literature. The phrase used in Google Scholar to find these articles was (`Procedural Content Generation' OR PCG) AND (game OR games), and the phrase used in Google was `Procedural Content Generation for games'. 
    \item We only included papers focusing on generating content, so articles similar to Osborn et al. (\citeyear{osborn2019your}), which is about evaluating the generated content and not generating it, are not part of this survey.
    \item We excluded research that focused solely on the behaviors of non-player characters (NPCs). While some behavior generation techniques may overlap with PCG, we feel that NPC behaviors are a separate topic that deserves its own analysis.
    \item While we focused on papers that talked about content generation in video games, we also mentioned articles that used PCG in tabletop games and board games.
    \item It is important to note that content generation has uses outside of designing and developing games for humans to experience. In addition to creating content in games meant for humans to play, PCG is also being utilized in scientific research to create game-like benchmarks and playgrounds for reinforcement learning and other forms of AI \cite{liu2021deep}. PCG is also being used in other formats, such as movies \cite{Massive_software} and art \cite{artstation}. In this paper, we only focus on game content generation. 
\end{enumerate}

\subsection{Novelty of the Work}
The differences between our work and previous surveys are that (i) we conduct a comprehensive review of both PCG methods and targeted content, (ii) at the time of writing, we are the first review paper to include LLMs as part of the methods used in PCG, and describe why LLMs are unique from all previous approaches, (iii) we add a section for combined methods and discuss why this category is important, and (iv) by analyzing papers published in the most recent 5 years in different conferences, we create a timeline for the methods used and find the current trend in academic research. 

\section{Content Types}

Almost everything, from sounds to the game narrative, can be generated nowadays, but the generated content for games can vary a lot depending on the algorithms used. For example, LLMs are mostly used to create game narratives \cite{huang2023create}, and GANs are considered a better solution to generate images \cite{kang2020image} and 2D levels \cite{abraham2023utilizing}.

There are several ways that we can categorize generated content. For example, it can be divided into online and offline generation. Online generation means that the content generation is performed during the runtime of the game, while offline generation means it is created during game development. Additionally, content can be categorized as necessary or optional \cite{togelius2011search}. For this survey, we use categories similar to the ones presented by Hendrikx et al. (\citeyear{hendrikx2013procedural}) with some modifications. We divide content created for games into five different categories, each consisting of multiple items. 

 \begin{enumerate}
     \item \textbf{Game bits:} This category consists of the smallest pieces (units) used in games. Any type of generated texture, sound, vegetation, structures and buildings, and object properties (e.g. can it interact with basic physics?) goes into this list. We also included art and images \cite{coutinho2022challenges} created by PCG algorithms in this category.
     \item \textbf{Game Space:} Game space is the physical environment that the game takes place and is created from game bits.  PCG algorithms are commonly used to generate maps and roads for games. Researchers have even tried to create entire game worlds with them \cite{prins2023storyworld}. 
     \item \textbf{Game Scenarios:} Game scenarios are parts of the game that are tied to the narrative. This includes generated conversations, stories, and quests. We include puzzles and interactive level elements in game scenarios because they are part of the scenarios that contribute to the story. This is especially true in the case of generated levels for side-scrolling 2D games \cite{kumaran2019generating}. Levels for music games, such as Guitar Hero or Dance Dance Revolution, can be seen as 2D levels as well \cite{liu2021deep}. 
     \item \textbf{Game Design:} This category consists of any mechanics or rules created for games (what can the player do and what are the goals?) It also includes generated system design, which we interpret as systems used in the games. Generating spawn points for first-person shooter games for game design purposes \cite{ballabio2019heuristics} is a perfect example of a generated design system. 
     \item \textbf{Derived content:} Derived content includes everything that is not essential to the game but can help the player better immerse in the game world. This category contains background NPC interactions, news found within the game, and chatter of different characters that are not part of the game’s story or a quest.
 \end{enumerate}

\begin{figure}
	\centering
\includegraphics[width=0.47\textwidth]{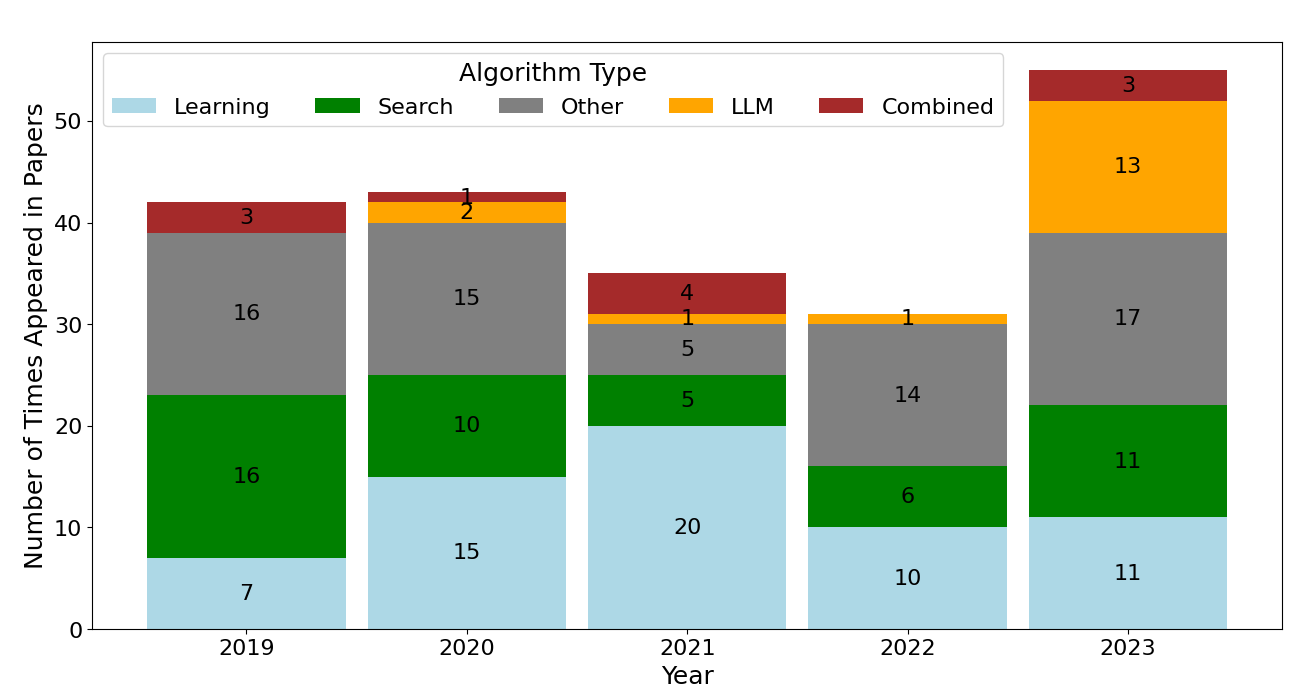}
        \caption{A timeline showing the types of algorithms appeared in PCG-related research papers during the most recent five years.}
	\label{fig:timeline2}
\end{figure}

\section{PCG Algorithms}
Due to the different types and roles of content in games, diverse PCG methods have been adapted for procedural content generation. In this section, we present different algorithms that exist and can be used to generate items. We categorized them into five main categories: Search-based, Machine Learning-based, Other, LLMs, and Combined Methods. The category called Combined Methods includes papers that use several methods in their study, either in parallel or in one integrated system.

It is worth mentioning that there were a few methods that we could not place under specific subgroups, so we decided to mention them at the start of each category.

\subsection{Search-Based Methods}

The term ‘search-based PCG’ was coined by Togelius in his paper on the taxonomy of PCG \cite{togelius2010search}. A search-based PCG refers to a special case of generate-and-test PCG. A generate-and-test PCG does not directly dish out content it generates but instead tests the content first using a test function. Depending on the test result, the PCG can either accept the content or reject it and create new content.

A search-based PCG is a test-and-generate PCG that satisfies two criteria \cite{prasetya2016search}. First, instead of simply accepting or rejecting content, the test function of a search-based PCG assigns a real value that measures the acceptability of the content. This value is often called fitness, and the function that produces it is called a fitness function. Second, in the creation of new, better content, a search-based PCG uses the previously rejected content as the creation base, and the new content is a slightly modified version of the old content. Search-based methods have been used to generate a variety of content, such as puzzles, race tracks, levels, terrains, and maps \cite{togelius2010search,prasetya2016search}.

In many cases, search-based algorithms use some form of evolutionary algorithm as the main search mechanism, as evolutionary computation has so far been the method of choice among search-based PCG practitioners. However, search-based PCG does not need to be married to evolutionary computation; other heuristic and stochastic search mechanisms are viable as well \cite{togelius2010search}. In this section, we discuss evolutionary algorithms, Wavefunction Collapse (WFC), Monte Carlo Tree Search (MCTS), simulated annealing, and particle swarm optimization. 
\begin{enumerate}
    \item \textbf{Evolutionary Algorithm}: In an evolutionary algorithm, a population of candidate content instances is held in memory. Each generation, these candidates are evaluated by the evaluation function and ranked. The worst candidates are discarded and replaced with copies of the good candidates, except that the copies have been randomly modified (i.e., mutated) and/or recombined \cite{togelius2010search}.
One of the most famous examples of evolutionary algorithms is genetic algorithms. Genetic algorithms are used in many instances \cite{baek2022toward, mitsis2020procedural, drageset2019optimising} to create playable levels and game bits. This algorithm can also be used to create more traditional types of games. For example, Botea and Bulitko (\citeyear{botea2023generating}) use this algorithm to create Romanian crossword puzzles. Other evolutionary algorithms are also used to create soundtracks for games \cite{makhmutov2019adaptive}, maps and game bits \cite{makhmutov2019adaptive}, roads \cite{song2019townsim}, and levels for board games \cite{gerhold2023computer}. 
\item \textbf{Planning Algorithms}: Planning, in general, is a problem-solving technique consisting of a planning problem (i.e., initial state and goal specification) and a planning domain (i.e., objects, predicates, and action operators). Given the input of a planning problem, a sound planner produces a solution or a plan, which is a sequence of actions that achieve all the specified goal conditions without any causal threats \cite{song2020intelligent}. In recent years, planning-based algorithms have often been used to generate stories in games because planning-based narrative generation is effective at producing stories with a logically sound flow of events \cite{siler2022open}.
    \item \textbf{Wave Function Collapse (WFC)}: WFC is a texture synthesis algorithm. Compared to earlier texture synthesis algorithms, WFC guarantees that the output contains only those NxN patterns that are present in the input. This makes WFC perfect for level generation in games and pixel art, and less suited for large full-color textures \cite{efros1999texture}. The WFC algorithm also supports constraints, allowing it to be easily combined with other generative algorithms or manual creation \cite{Mxgmn}. 
    \item \textbf{Monte Carlo Tree Search (MCTS)}: Monte Carlo Tree Search (MCTS) is a heuristic search algorithm that involves searching combinatorial spaces represented by trees. In such trees, nodes denote states whereas edges denote transitions (actions) from one state to another \cite{swiechowski2023monte}. 
    \item \textbf{Simulated Annealing:} Simulated annealing is a process where a setup is randomly tweaked and compared to the previous setup using the cost function. If it is better, the new setup is kept. Otherwise, the decision to keep the new setup is made randomly, with the probability of keeping the old setup proportional to how much better it is \cite{russell2016artificial}.
    \item \textbf{Particle Swarm Optimization (PSO)}: PSO is a general-purpose optimization technique developed by Eberhart and Kennedy (\citeyear{kennedy1995particle}). This technique was inspired by the concept of swarms in nature, such as bird flocking, fish schooling, or insect swarming. The idea is that individual members of the swarm can profit from the discoveries and previous experiences of all other members of the swarm during the search for the optimum solution \cite{duro2008particle}. 
\end{enumerate}

\subsection{Machine Learning-Based Methods}

Machine learning methods have gained a lot of popularity during the last decade \cite{summerville2018procedural}. Research on neural networks under the name deep learning has precipitated a massive increase in the capabilities and application of methods for learning models from big data \cite{schmidhuber2015deep, Goodfellow-et-al-2016}. They have provided us with new ways for generating audio, images, 3D objects, network layouts, and other content types \cite{goodfellow2014generative} across a range of domains, including games. For example, long short-term memory (LSTMs) are mostly used for time-dependent sequential data (e.g., action sequences, agent paths, charts for rhythm) and language models \cite{risi2020increasing}, while generative adversarial networks (GANs) have been applied to generate artifacts, such as images, music, and speech \cite{goodfellow2014generative}. Outside of content generation, machine learning algorithms have been widely used in testing generated maps using other algorithms \cite{liu2021deep}. Many other machine learning methods can also be utilized in a generative role, including n-grams, Markov models, autoencoders, and others \cite{boulanger2012modeling,fine1998hierarchical,gregor2015draw,summerville2018procedural}. 

In this section, we talk about different machine learning methods such as Autoencoders (including deep variational autoencoders), Recurrent Neural Networks (including LSTMs), Generative Adversarial Networks (GANs), Markov Models, Reinforcement Learning methods, and Transformers.

The most recent entry, LLMs, are gaining a lot of attention among researchers. While state-of-the-art LLMs use transformers as their underlying architecture, their model-as-a-service (MaaS) nature puts them into a different category.

It is worth mentioning that not all articles regarding deep learning methods use complex neural network models. Merino et al. (\citeyear{merino2023five}) use neural networks to create sprites and game maps for 2D games.  Bhaumik et al. (\citeyear{bhaumik2023lode}) and Kumaran et al. (\citeyear{kumaran2019generating}) use neural network models to generate maps and levels for 2D games.

\begin{enumerate}
    \item \textbf{Simple Neural Networks}: Some works on PCG algorithms rely on the use of relatively simple neural networks. For example, papers such as Chen etl al. (\citeyear{chen2020image}), take images as inputs and output game levels using neural networks.
    \item \textbf{Recurrent Neural Networks \& LSTMs}: A recurrent neural network (RNN) is a type of artificial neural network that uses sequential data or time series data. They are distinguished by their “memory” as they take information from prior inputs to influence the current input and output \cite{medsker2001recurrent}. LSTMs are similar neural networks introduced to eliminate the vanishing gradient problem in RNNs. LSTMs work to solve that problem by introducing additional nodes that act as a memory mechanism, telling the network when to remember and when to forget. As mentioned before, LSTMs and RNNs are mostly used to create sequential data. Summerville and Mateas (\citeyear{summerville2016super}) deals with Mario levels as a sequence of data and combines Markov Chains and LSTMs to create new levels. 
    \item \textbf{Generative Adversarial Networks (GANs)}: GANs are very popular for content generation purposes. They usually consist of two networks, a generator and a discriminator, that are trained iteratively to allow the generator to create more realistic content while the discriminator gets better at distinguishing generated content from real data \cite{goodfellow2014generative,liu2021deep}.

GANs are perfect for generating content represented by pixel-based images or 2D arrays of tiles, such as levels, maps, landscapes, and sprites. By reviewing the gathered papers, it can be seen that in the work using GANs for level generation, game levels are tackled as images only during training, while the constraints for validating levels are not considered at all. 
    \item \textbf{Markov Models}: The Markov chain algorithm is a typical constructive method. In this approach, content is generated on-the-fly \cite{Kernighan_Pike_2006} according to the conditional probabilities of the next state in a sequence based on the current state. This state can incorporate multiple previous states via the construction of n-grams. An n-gram model (or n-gram for short) is simply an n-dimensional array where the probability of each state is determined by the n states that precede it \cite{summerville2018procedural}. This is the simplest form of Markov chains, also called 1D Markov chains.

There are also multidimensional Markov chains (MdMCs) \cite{ching2007multi}, where the state represents a surrounding neighborhood and not just a single linear dimension. A multidimensional Markov chain differs from a standard Markov chain in that it allows for dependencies in multiple directions and from multiple states, whereas a standard Markov chain only allows for dependence on the previous state alone \cite{summerville2018procedural}. In addition to their standard MdMC approach, Snodgrass and Ontañón developed a Markov random field (MRF) approach \cite{goodfellow2014generative} that performed better than the standard MdMC model in Kid Icarus, a domain where platform placement is pivotal to playability \cite{snodgrass2016learning}. This method has generated novel but recognizable game names \cite{browne2008automatic}, natural language conversations, poetry, jazz improvisation \cite{pachet2004beyond}, and content in a variety of other creative domains \cite{bentley1999three}. 
\item \textbf{Autoencoders}: An autoencoder is a type of neural network architecture designed to efficiently compress (encode) input data down to its essential features, then reconstruct (decode) the original input from this compressed representation. One of the famous autoencoders used for PCG is variational autoencoders (VAEs). VAEs are generative models that learn compressed representations of their training data as probability distributions, which are used to generate new sample data by creating variations of those learned representations \cite{ng2011sparse}. These algorithms are mostly used for generating 2D maps and levels. 

Snodgrass and Sarkar (\citeyear{snodgrass2020multi}) use VAEs to generate level structures and a search-based approach to blend details from various platformers, while Sarkar et al. \cite{sarkar2020exploring} directly trains VAEs on levels from several platforming games and interpolates the latent vectors between domains for blending \cite{liu2021deep}. Sometimes, trained autoencoders may be used to repair unplayable levels. Davoodi et al. (\citeyear{davoodi2022approach}) train an autoencoder to repair manually designed levels for different games by re-iterating it over the decoder while using a trained discriminator from a GAN model to determine the stopping criteria \cite{liu2021deep}. 
\item \textbf{Reinforcement Learning}: RL problems involve learning how to map situations to actions that maximize a numerical reward signal \cite{sutton2018reinforcement}. Most articles that use RL develop agents to play generated levels, which indirectly serve as content evaluators.

To generate content using RL, the generation task is usually transformed into a Markov decision process (MDP), where a model is trained to iteratively select actions that would maximize expected future content quality. This transformation is not an easy task, and there is no standard way of handling it. Most RL PCG approaches require an adaptation of the input to be used during generation \cite{liu2021deep}. There are also some interesting cases where RL can be used in the absence of data, provided a system for the learning agent to interact with can be set up \cite{barriga2019short}. 
\item \textbf{Transformers}: Transformers are a type of neural network architecture that transform an input sequence into an output sequence by learning context and tracking relationships between sequence components \cite{khan2022transformers}. They rely on a self-attention mechanism \cite{vaswani2017attention}, which facilitates the capture of intricate semantic relationships within high-dimensional feature spaces. For example, PCGPT \cite{mohaghegh2023pcgpt} used transformers to iteratively generate complex and diverse game maps in the Sokoban game.

\subsection{Other Methods}

There are some methods that do not belong in the previous categories. We added this list for frequently used methods in PCG that we could not fit in any other category. It includes pseudo-random number generators (PRNGs), generative grammars, generative graphs, and fractals. 

It is worth mentioning that these categories are not the only means of generating content for games. There are some articles about innovative ways of PCG. For example, Wootton (\citeyear{wootton2020procedural}) uses quantum blur effects to create maps and levels for various games. Wootton (\citeyear{wootton2020quantum}) also uses quantum computation combined with graphs to generate maps. 
     \begin{enumerate}
         \item \textbf{Pseudo-random Number Generator (PRNG):} One of the simplest and earliest approaches to procedural game content generation is based on pseudo-random number generation (PRNG). PRNG is an algorithm for generating a sequence of numbers that approximates the properties of random numbers \cite{barker2007sp}. A PRNG algorithm consists of three parts: the seed, which is the initial value, the formula, which converts the seed to output, and the distribution, which is the variance of the results \cite{matsumoto2006pseudorandom}.

PRNG was first used as a data compression method because the generated sequence, while appearing random, can be reproduced if the same seed and algorithm are used. Combined with other methods, PRNG-based techniques can be used to generate buildings, textures, and items \cite{hendrikx2013procedural}. One of the most famous forms of PRNGs is noise functions. Perlin noise and other noise functions are commonly used for texture generation. Noise-generated textures can be mapped easily on complex objects, unlike raster 2D images. The implementation of noise is relatively simple and is present in many software shaders and hardware graphics cards, such as NVIDIA’s \cite{hendrikx2013procedural}. One of the main drawbacks of this method is that images generated randomly pixel-by-pixel have no meaningful structure. Many procedural techniques address this issue by finding the balance between iterative generation and random generation.

         \item \textbf{Generative Grammars:} Generative grammars, stemming from Noam Chomsky’s study of languages in the 1960s \cite{hendrikx2013procedural}, are sequences of words (or “sentences”) with rules regarding how and when to replace some words with other words. They can be used to create correct objects from elements encoded as letters/words. L-systems, split grammars, wall grammars, and shape grammars, which have been used for plant generation, linear map dungeon or story generation, and quest generation, are all part of generative grammars. The downside of using this method is that it is linear, so it cannot be used generate a non-linear story. The possible solution is using generative graphs, which we discuss in the following section.
         \item \textbf{Generative Graphs:} A graph is a set of vertices connected by edges. Generative graphs are used as a solution for the linearity problem of generative grammars. One well-known type of generative grammar is graph grammars. A graph grammar is a set of rules that modify a graph. It is a framework introduced decades ago \cite{PFALTZ1972193,ehrig1973graph}.

One of the challenges of using graph grammars was the need for an expert to design the rules, but now, some researchers \cite{merrell2023example} have worked on creating the grammar automatically. The content generated using generative graphs is almost similar to that generated by generative grammars, only they do not have to be linear. One of the most famous products of graph grammars is SpeedTree \cite{ehrig1973graph}, a well-known set of tools to generate trees in the entertainment industry \cite{Columbia_Metropolitan_Magazine_2019}.
         \item \textbf{Fractals:} The term "fractal" was coined and popularized by Benoit B. Mandelbrot \cite{mandelbrot1985self}. It describes a broad set of shapes characterized by non-integer dimension or an interesting mismatch of dimension (Hausdorff dimension strictly exceeding topological dimension), and detail at all scales or self-similarity. Fractals are frequently used in procedural content generation because self-similarity seems to mimic natural processes such as erosion and plant growth. The subdivision method also maps well onto level of detail implementations, allowing an 'infinite' amount of detail to be included by recursively subdividing the detail shown as the viewpoint moves closer to the fractal object \cite{cristea2015fractal}.

     \end{enumerate}

\end{enumerate}

\subsection{Large Language Models}

In the most recent year, there has been an explosion of research on the applications of LLMs, and PCG is no exception. We found 17 papers during these years using LLMs as part of their pipeline (with 3 other papers using combined methods that integrated LLMs). Referring back to Figure 1, we see that in 2023, the use of LLMs drastically increased compared to previous years (13 papers and 2 other papers with combined methods integrating LLMs). This coincides with the release of ChatGPT and its underlying model, GPT-3. These models are used to create narratives, NPC chatter, and even mechanics. A famous example is Dungeon 2, a text adventure game \cite{schrum2020cppn2gan}. In this game, players can type in any command and the system can respond to it reasonably well, creating the first never-ending text adventure. The system is built on OpenAI’s GPT-2 language model \cite{volz2018evolving}, which was further fine-tuned on a number of text adventure stories. 1001 Nights is also another project that uses GPT as one of the main mechanics of its gameplay \cite{sun2023language}. Language models are also used in role-playing board games as an assistant for the game masters \cite{zhu2023calypso}. LLMs can also be used to create game levels. SCENECRAFT \cite{kumaran2023scenecraft} is a framework that transforms high-level natural language instructions from authors into dynamic game scenes that include NPC interactions, dialogue, emotions, and gestures. 
The research by Todd et al. (\citeyear{todd2023level}), Nasir and Togelius (\citeyear{nasir2023practical}), and Sudhakaran et al. (\citeyear{sudhakaran2024mariogpt}) also focus on using LLMs to create levels for games. For example, the latter uses MarioGPT, a fine-tuned GPT-2 model designed to generate Super Mario Bros levels based on textual prompts.

While current LLMs use transformers as their underlying model, with the publication of GPT-3, a new form of service has emerged. ``Model as a service" (MaaS) involves deploying a model on a cloud-based infrastructure and offering its functionalities through APIs or web interfaces. These commercialized models, such as GPT-3, are often no longer open for researchers to explore and study their underlying architecture, creating a black box that is only known to select few. Our review shows that research published on using GPT-3 and its successors is no longer about training a new model, but about best ways to use an existing pre-trained model that is restricted by a private entity in many ways. While open source LLMs do exist, in our review, the vast majority of LLM usage was still with GPT-based LLMs.

\begin{figure*}[!htb]
	\centering
	\includegraphics[width=1\textwidth]{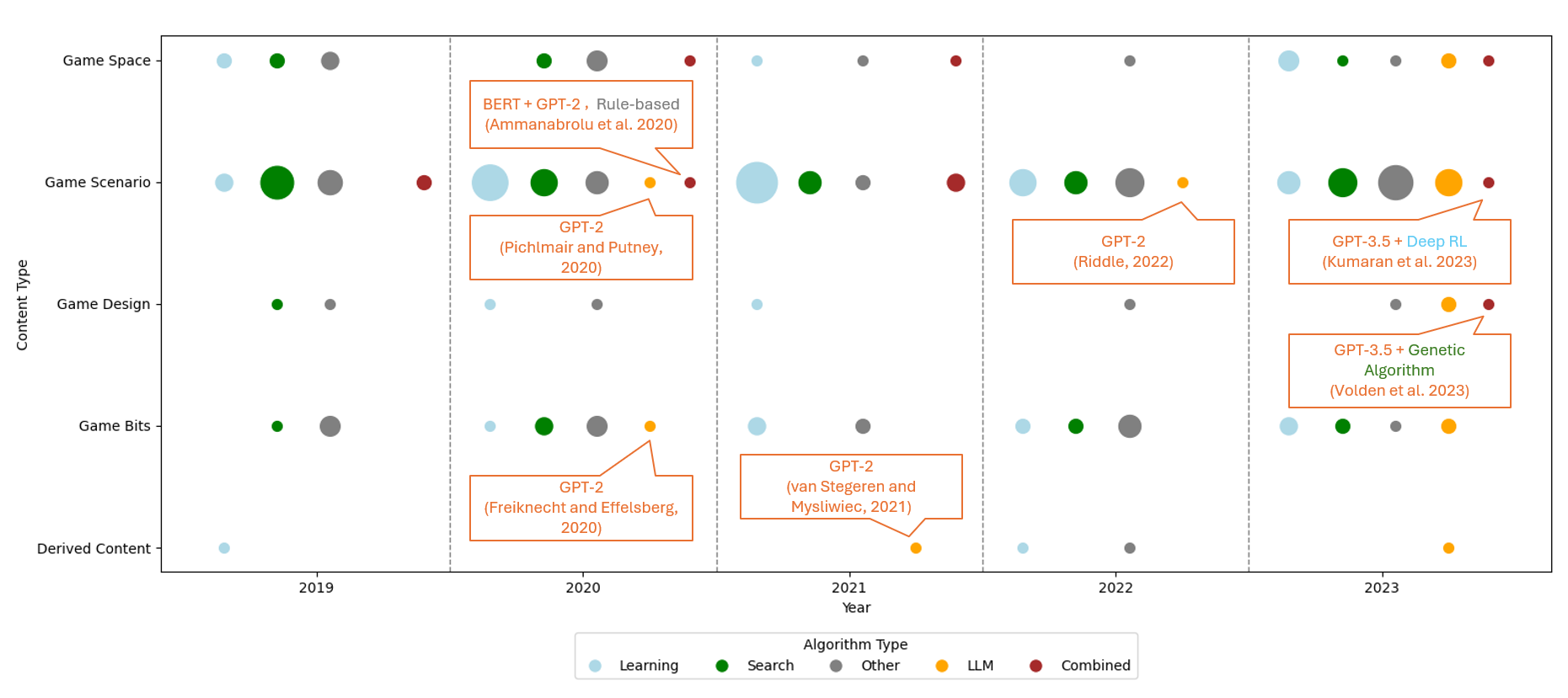}
        \caption{A timeline breaking down research published 2019-2023 in select game-related conferences on the topic of PCG, sorted by targeted content type. A few note-worthy LLM-based works are pointed out.}
	\label{fig:timeline}
\end{figure*}

\subsection{Combined Methods}

So far, we have discussed four different methods used for generating content in games. However, these methods can be combined. Many researchers have worked on using evolutionary algorithms with machine learning methods. One of the most popular examples of combining methods is the Latent Variable approach \cite{bontrager2018deepmasterprints}, which combines unsupervised learning in the form of a GAN or VAE with evolutionary computation to search for content in the learned space of a GAN/VAE. In the context of games, this approach has been employed to generate 2D game levels like Super Mario Bros and Zelda levels \cite{schrum2020cppn2gan,volz2018evolving}. RL is also used in combination with many different algorithms. For instance, Kumaran et al. (\citeyear{kumaran2023end}) creates game levels through a natural language interface and evaluates the levels using RL. Instead of relying solely on one method, combining different methods can be very effective in generating new content. RL models and search-based algorithms are effective tools to repair generated content (especially levels and structures) created by other algorithms (like GANs and LSTMs). More recently, combined methods integrating LLMs have also begun to emerge \cite{kumaran2023end, volden2023procedurally}. While PCG research tends to focus on using existing methods in innovative ways, these combined methods form a rather emerging approach in solving complex problems.

\section{Analysis}

In this section, we discuss the information obtained from reviewing selected papers. First, we talk about the trends in the papers. Then, we discuss the gaps in current works, and finally, we outline future research directions based on current trends. 

\subsection{Trends}

Figure 2 shows an overview of 207 published papers using different methods for the period of 2019-2023, broken down by content type. Each row represents one type of content, such as the aforementioned Game Space. The dots represent papers in that particular year - the larger the dot, the more papers we found. The dots are color-coded by its algorithm type. We also gathered information about the 629 authors who worked on these papers which is shown in Figure 3 based on the number of co-authored papers.

By looking at Figure 2 which is extracted using collected articles, we notice that level generation is by far the most researched topic in PCG. There are 102 papers, almost half of the selected papers, focusing on level generation. Except for a few works such as Baek et al. (\citeyear{baek2022toward}), most of these papers focus on 2D level generation. This seems reasonable as creating a 3D level is a far more complex task than creating a 2D level.
We also found that 2D platformers are a popular genre in PCG, with \textit{Super Mario Bros.} being the most used environment, featured in 18 papers. 

From Figure 2, it is evident that machine learning-based algorithms have gained significant interest, and the trend of using them in PCG continues to grow. Liapis et al. (\citeyear{liapis202010}) mentions the consistent inclusion of vision papers and the increased use of machine learning in games, either on its own or in conjunction with PCG. Our data shows that the trend has not slowed down in recent years. The emergence of ChatGPT \cite{openai2024chatgpt} and the explosion of LLMs like BERT and GPT in recent years have also affected content generation in games. Many of the papers using LLMs were published in the last year. It is interesting to note that we found only five articles published before 2023 that used any type of LLMs, and all five used GPT-2 (which is an open-source model), with one also using BERT (another open-source model) \cite{riddle2022hybrid, van2021fine, ammanabrolu2020bringing, pichlmair2020procedural, freiknecht2020procedural}.  2023 was the year when many LLM-based research emerged, but also when people moved on from open-source models to closed-source, mostly GPT-based services. Among the works, two caught our attention. These are combined methods with LLMs. Kumaran et al. (\citeyear{kumaran2023end}) combined GPT-3.5 with the use of deep reinforcement learning for level generation and selection, while Volden et al. (\citeyear{volden2023procedurally}) put together LLMs with a genetic algorithm to generate rules for difficulty adaptation in a serious game. We believe that these combined methods will show versatility in many situations, and this is an open research field with a lot of potential.

\begin{figure*}[!htb]
	\centering
	\includegraphics[width=0.86\textwidth]{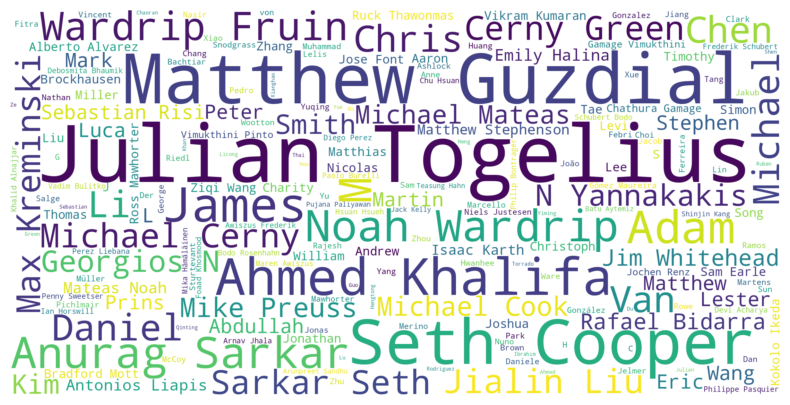}
        \caption{A word cloud of all authors in the published papers at the selected conferences during the 5 years. The size is proportionate to the number of co-authored papers.}
	\label{fig:wordcloud}
\end{figure*}

\subsection{Gaps in Research}
As mentioned, most of the research in level generation focuses on 2D level generation. 3D games are very popular in the games industry, and level generation for 3D games is a topic that needs further research and exploration.

Based on our findings, despite having innovative ideas in the selected papers, many do not follow up on those ideas, with some rare exceptions such as 1001 Nights \cite{sun2023language}. This gap between academic work and the games industry requires more discussion. One possible solution could be creating a playable demo of a game with generated content (if feasible). This approach could increase the chances of finding investors for the research. Of course, we cannot ignore the ethical considerations when using tools such as LLMs for generating game content \cite{melhart2023ethics}.

\subsection{Possible Research Directions}
Using the data we gathered, we believe that combined algorithms, especially deep learning algorithms and LLMs, will be a popular topic in the future. Instead of trying to generate whole items or levels, researchers could focus on repairing incorrect or improper content, reducing the effort needed to create a finished product.

LLMs are still fresh, and there is much unknown potential in them that can be used in PCG. They can serve as AI assistants for game designers, story and quest generators, and even level generators that produce levels as a sequence of words or sentences. Therefore, more research on the applications of LLMs could prove beneficial, especially open-source alternatives to closed-source commercial MaaS.

\section{Conclusion}
In this survey, we discussed different categories of PCG algorithms and the content that can be generated for games. We categorized PCG algorithms into five sections: search-based methods, machine learning-based methods, other methods, LLMs (MaaS), and combined methods. We also selected PCG related papers published between 2019-2023 and analyzed them to identify trends, gaps in current work, and possible future research directions.

The increasing use of LLMs for generative tasks is a recent development that was unheard of until about five years ago. Combining deep learning methods, such as GANs and RL, with other algorithms in content generation is another interesting trend that has been gaining more attraction recently. Both of these trends are built on advancements in deep learning, which have made machine learning methods effective for completely new classes of problems.

Although a variety of generated content types (e.g., levels, sound, maps, textures) have been investigated, each subcategory can be further explored to gain more insights. In-depth research on the best-suited algorithms for generating each type of content would also be beneficial. Additionally, a study dedicated to evaluation algorithms used to test generated content could be a valuable addition to this field.

\section{Acknowledgments}

This research was supported by the Natural Sciences and Engineering Research Council of Canada (NSERC) Discovery Grant. We thank members of the Serious Games Research Group and the anonymous reviewers for their feedback.

\bibliography{aaai24}

\end{document}